\documentclass[letterpaper, 10 pt, conference]{ieeeconf}  % Comment this line out if you need a4paper

\IEEEoverridecommandlockouts                              % This command is only needed if 
\usepackage{graphics} % for pdf, bitmapped graphics files
\usepackage{booktabs} 
\usepackage{multirow}
\usepackage{array}
\usepackage{amsmath}
\usepackage[table]{xcolor}
\usepackage{multirow}
\usepackage{graphicx}
\usepackage{hyperref}
\usepackage{amsmath,amssymb}
\title{\LARGE \bf
DexTouch-WM: Learning Action-Conditioned Tactile World Models from Human Touch for Dexterous Robot Manipulation
}

\newcommand{\method}{DexTouch-WM}
\newcommand{\shorttask}[1]{\shortstack{#1}}
\author{
Yan Qin$^{1,2,*}$, Yue Chen$^{2,3,*}$, Wenwei Lin$^{2,5,*}$, Shujia Liu$^{5,*}$, Chuqiao Lyu$^{2,5}$, Kailun Su$^{2,5}$,\\
Weiyang Jin$^{4}$, Chenze Yu$^{2,5}$, Ping Luo$^{4}$, Wenbo Ding$^{2,5,\dagger}$, Tianxing Chen$^{2,4, \dagger}$ and Renjing Xu$^{1,\dagger}$
\thanks{*Equal contribution and shared first authorship.}%
\thanks{$\dagger$Corresponding author: Renjing Xu.}%
\thanks{$^{1}$HKUST (GZ); $^{2}$Xspark AI; $^{3}$PKU; $^{4}$HKU; $^{5}$THU.}%
}

\begin{document}

% Full-width teaser under title/authors and above the abstract.
% Must be issued before \maketitle. Conference mode locks this command
% unless \IEEEoverridecommandlockouts is already set (it is, above).
\makeatletter
\IEEEaftertitletext{%
% \vspace{0.5\baselineskip}%
\begin{minipage}{\textwidth}%
\centering
\includegraphics[width=\textwidth]{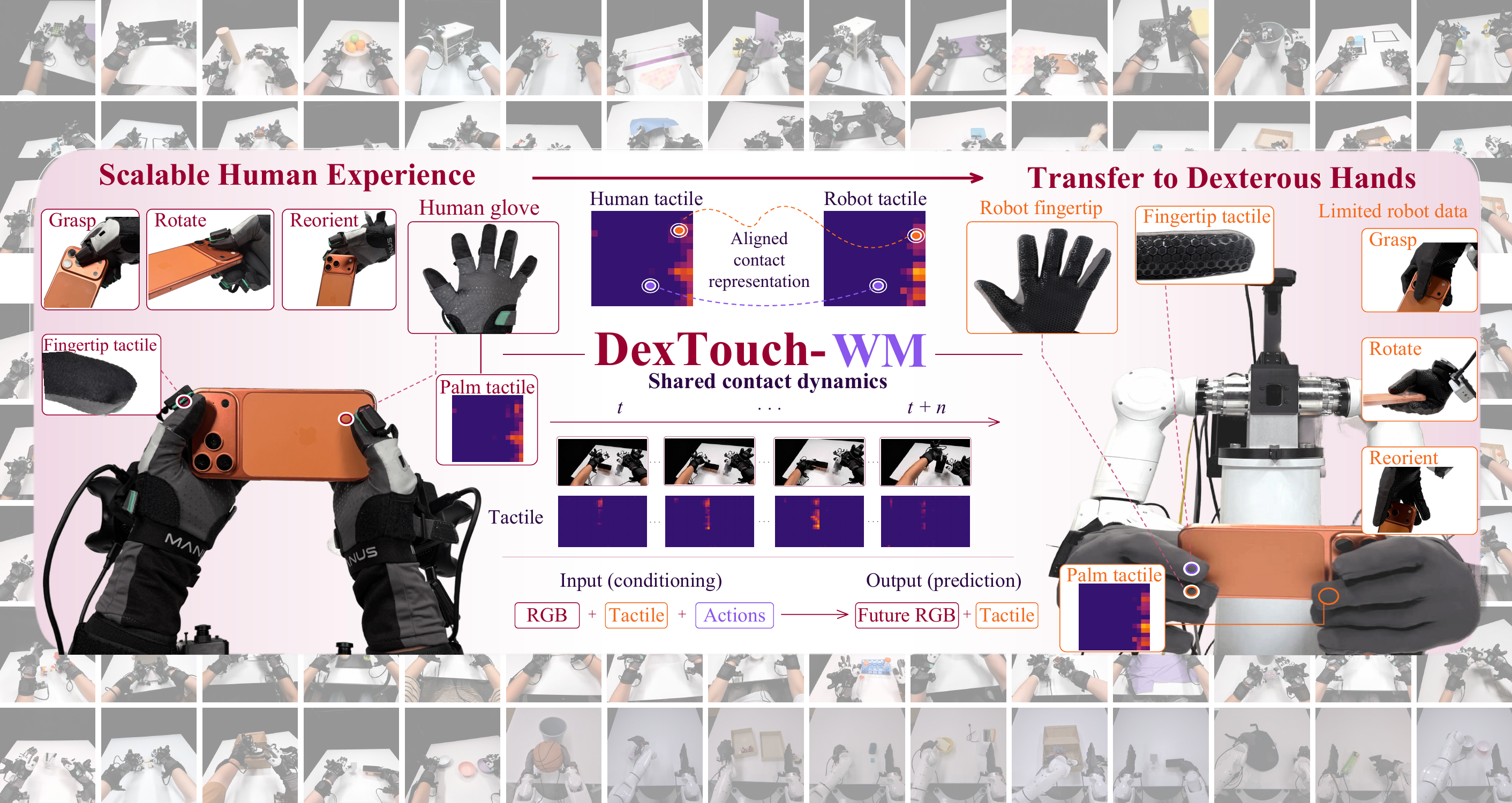}%
\def\@captype{figure}%
\caption{Overview of \method{}. Given initial observations and an action sequence, \method{} jointly predicts future video and tactile maps. These human trajectories provide a scalable contact-dynamics prior for subsequent robot-domain adaptation.}%
\label{fig:teaser}%
\end{minipage}%
}
\makeatother

\maketitle
\thispagestyle{empty}
\pagestyle{empty}

\begin{abstract}
Learning predictive models of contact-rich dexterous manipulation requires
dense tactile interaction, but such data are costly to scale on real robots
and are tied to embodiment-specific sensors. We introduce \method{}, an action-conditioned world model that learns from scalable human touch to jointly predict future RGB observations and bilateral tactile dynamics. Our insight is that human and robot manipulation share transferable
contact dynamics when their tactile observations and action spaces are made
compatible. We deploy flexible piezoresistive arrays with a
shared sensing layout on both human and dexterous robot hands, and retarget
human motion into the robot action space so that human
interaction can supervise the same dynamics model used for real-robot
prediction. \method{} couples a
pretrained video expert with a lightweight tactile expert using
anatomy-aware tactile tokens and aligned action conditioning. In human-to-robot scaling
experiments, we keep real-robot supervision fixed while increasing human
interaction from 0 to 100\,h, and observe substantial improvements in
held-out robot-domain visual, geometric, and contact prediction despite
disjoint human and robot task sets. Beyond prediction, we evaluate
the world models as surrogate environments for policy evaluation and
as generators of synthetic trajectories for real-robot policy
learning, showing that scalable human interaction provides a
complementary data axis for learning dexterous robot world models.
\end{abstract}
\section{Introduction}
\label{sec:introduction}

Contact-rich dexterous manipulation is governed by physical interaction states that are not directly observable from images. Visually similar motions can correspond to different contact pressure, slip, or incipient jamming, particularly when the hand occludes the object. Tactile is critical for perceiving and controlling such interactions~\cite{huang2024vitac,xue2025reactive}. Action-conditioned world models predict how observations evolve under candidate action sequences and are increasingly used for policy evaluation~\cite{worldgym,gigaworld}, learned simulation~\cite{interactiveworldsim}, and synthetic data generation~\cite{worldsample}. Models that predict only video cannot represent the contact dynamics that determine manipulation outcomes. Recent visual-tactile world models address this limitation by jointly predicting visual and tactile observations and have shown benefits in physical fidelity and downstream manipulation~\cite{vtwm,vitacworld,dreamtac,vtwam}.

Data collection remains a major obstacle to extending these models to dexterous manipulation. Most existing approaches collect tactile trajectories through robot teleoperation, coupling each dataset to a specific embodiment and sensor configuration~\cite{vtwm,vitacworld}. Acquiring additional data therefore requires continued operation of the same platform, so available tactile datasets remain small and are often limited to parallel-jaw grippers or isolated fingertip sensors.

Our central idea is to decouple tactile data collection from the robot embodiment through wearable sensing. Lightweight piezoresistive gloves capture dense bilateral full-hand pressure during natural human interaction, and the same sensing layout can be deployed on robotic dexterous hands~\cite{dexumi,tactidex}, giving human and robot observations a common tactile representation. Hardware correspondence alone, however, does not resolve the modeling challenges. Full-hand taxels are distributed across spatially disconnected fingertip and palm regions with distinct anatomical roles, so a regular image representation is poorly aligned with their physical topology. Egocentric bimanual actions further combine articulated hand motion with camera motion and must be temporally aligned with visual and tactile streams operating at different rates.

We introduce \method{}, an action-conditioned tactile world model for full-hand dexterous interaction. \method{} couples a pretrained video expert~\cite{wan} with a lightweight tactile expert through Mixture-of-Transformers blocks~\cite{mot}, enabling joint multimodal interaction while retaining modality-specific computation. An anatomy-aware tactile tokenizer encodes fingertip and palm regions according to their physical sensing topology. Tactile dynamics are represented as residual latents anchored to the initial tactile observation, focusing prediction on action-induced contact changes rather than static pressure backgrounds. Dual-rate action conditioning injects hand and camera motion at the native temporal resolution of each modality, while a shared robot-centric pose representation maps human and robot motion into a common conditioning space.

We train \method{} on 100 hours of egocentric human interaction spanning 50 tasks, together with 5 hours of real-robot interaction. Holding robot supervision fixed while scaling human data improves held-out robot-domain visual, geometric, and contact prediction. We further evaluate the learned world models as surrogate environments for policy evaluation and as generators of synthetic trajectories for downstream real-robot policy learning.

\textbf{Our contributions are threefold:}
\begin{itemize}
\item We introduce an action-conditioned tactile world model that jointly predicts future video and dense bilateral full-hand tactile observations for dexterous interaction.

\item We develop a structured approach to tactile dynamics modeling that combines anatomy-aware tactile tokenization, initial-observation-anchored residual latents, and temporally aligned dual-rate action conditioning.

\item We construct a 100-hour dataset of bimanual human interaction and a shared human--robot sensing and action interface, showing that human data scales robot-domain prediction and supports policy evaluation and synthetic data generation.

\end{itemize}

\section{RELATED WORK}

\textbf{Tactile learning and human data.}
Tactile-aware policies combine visual context with contact feedback for fine manipulation~\cite{huang2024vitac,xue2025reactive}, predict tactile features for dexterous control~\cite{heng2026vitacformer}, or ground actions in evolving multi-point contacts~\cite{xu2026contactgrounded}. FTP-1 transfers tactile policy knowledge across sensors and embodiments~\cite{yuan2026ftp}, while TactAlign aligns human and robot touch for policy transfer~\cite{wi2026tactalign}. These methods treat touch as a control or alignment signal, rather than as a prediction target for how contact evolves under future motion. Human-collected visuo-tactile data, from Touch and Go~\cite{yang2022touch} to wearable or portable systems such as OpenTouch~\cite{opentouch}, DexUMI~\cite{dexumi}, FreeTacMan~\cite{freetacman}, DexViTac~\cite{dexvitac}, and Touch in the Wild~\cite{touchinthewild}, reduce reliance on robot-operated collection, typically to support imitation or representation learning. Our focus is the transfer of action-conditioned interaction dynamics: shared sensing and retargeted actions allow human trajectories to supervise the same world model as robot data.

\textbf{Visual--tactile world models.}
SPOTS predicts coupled visual and tactile observations~\cite{mandil2026multimodal}, while VT-WM learns action-conditioned dynamics for visuo-tactile interaction~\cite{vtwm}. OmniVTA~\cite{zheng2026omnivta}, FeelWorld~\cite{ma2026feelworld}, and ViTacWorld~\cite{vitacworld} further study contact-aware prediction, planning, or scalable rollouts. Dream-Tac and VT-WAM couple these predictions with action generation~\cite{dreamtac,vtwam}, and TacForeSight uses force-guided tactile forecasts as contact priors~\cite{zang2026tacforesight}. TouchWorld uses human tactile pretraining and predictive subgoals for dexterous manipulation~\cite{zhou2026touchworld}. Thus, using human touch and predicting tactile signals are established directions. \method{} studies their combination through distributed full-hand representations, a shared human--robot interface, and a controlled scaling protocol that holds robot supervision fixed. We thereby test a central hypothesis: shared contact dynamics allow scalable human physical interaction to scale dexterous-hand world models.

\begin{figure*}[t]
\centering
\includegraphics[width=\textwidth]{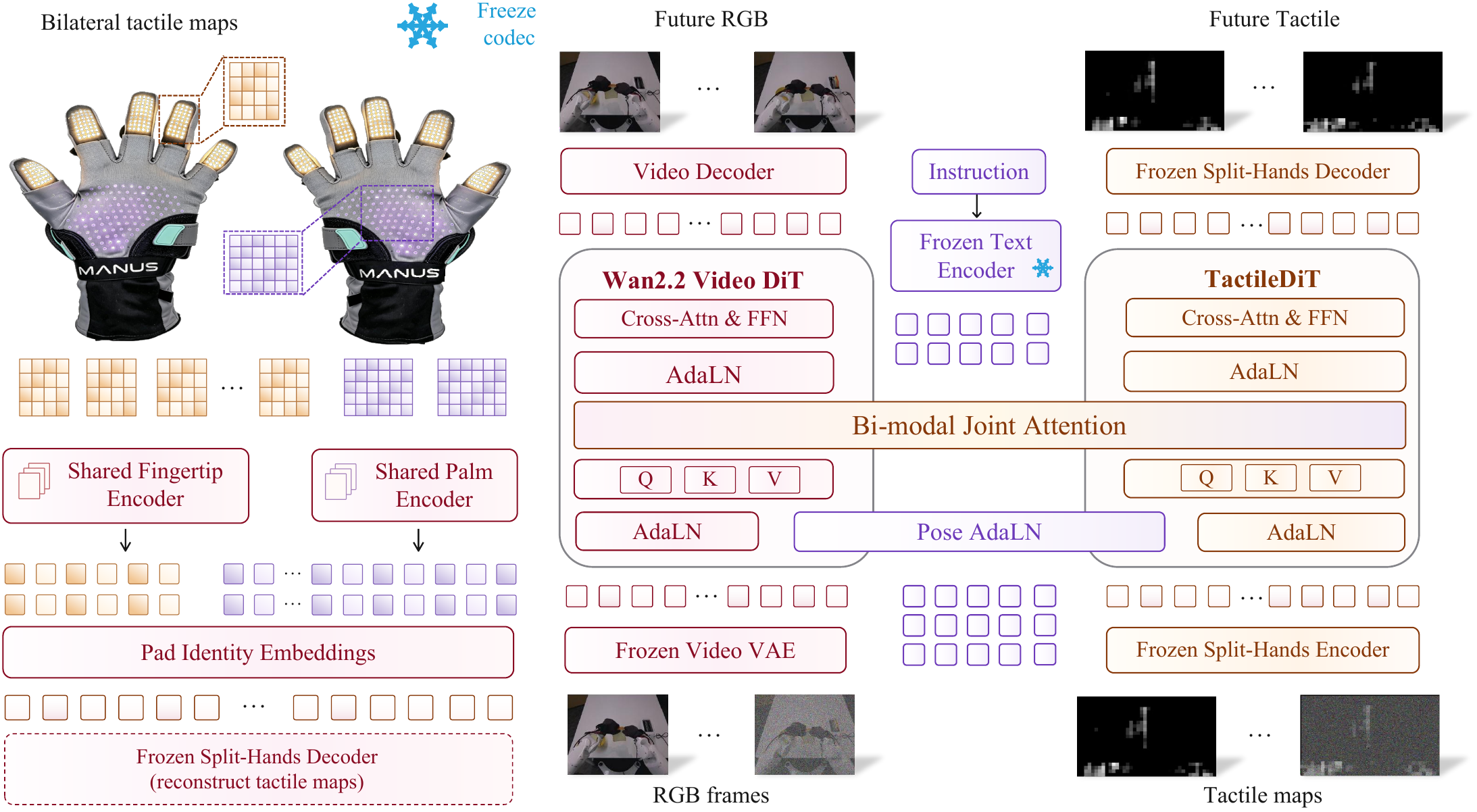}
\caption{\textbf{\method{} architecture.} Left: shared fingertip and palm encoders tokenize bilateral tactile maps while pad identity embeddings preserve anatomy. Right: a pretrained video expert and a tactile expert exchange information through bi-modal joint attention, retaining modality-specific AdaLN and feed-forward pathways. Instruction features and pose AdaLN condition generation. Frozen visual and tactile codecs encode observations and decode future RGB and pressure maps.}
\label{fig:pipeline}
\end{figure*}

\section{Method}
\label{sec:method}
\subsection{Problem Formulation}
\label{sec:method_formulation}
% Given an initial RGB observation $\mathbf{v}_0$, bilateral tactile map $\mathbf{x}_0$, proprioceptive state $\mathbf{s}_0$, language instruction $\ell$, and future actions $\mathbf{a}_{1:T}$, we learn
% \begin{equation}
%  p_\theta(\mathbf{v}_{1:T},\mathbf{x}_{1:T}\mid
%  \mathbf{v}_0,\mathbf{x}_0,\mathbf{s}_0,\mathbf{a}_{1:T},\ell).
%  \label{eq:problem}
% \end{equation}
% Human demonstrations contain tracked wrist poses and 25-keypoint hand skeletons. We retarget them into bilateral robot end-effector poses and 20-DoF hand configurations before training (Sec.~\ref{sec:human_robot_alignment}). This supplies a common action parameterization for both domains.
Given an initial RGB observation $\mathbf{v}_0$, bilateral tactile map
$\mathbf{x}_0$, language instruction
$\ell$, and future pose sequence $\mathbf{a}_{1:T}$, we learn
\begin{equation}
p_\theta(\mathbf{v}_{1:T},\mathbf{x}_{1:T}\mid
\mathbf{v}_0,\mathbf{x}_0,\mathbf{a}_{1:T},\ell).
\label{eq:problem}
\end{equation}
Human demonstrations contain tracked wrist poses, 25-keypoint hand
skeletons, and head-camera motion. We retarget human hand motion to the robot embodiment(Sec.~\ref{sec:human_robot_alignment}) and represent both domains using
the same 67-D pose representation, consisting of bilateral wrist motion,
bilateral 20-DoF hand configurations, and head-camera motion.

\subsection{Action-Conditioned Tactile World Model}
\label{sec:method_model}
As shown in Fig.~\ref{fig:pipeline}, \method{} combines a visual expert initialized from Wan2.2-TI2V-5B and its video VAE~\cite{wan} with a lightweight tactile expert. Following Mixture-of-Transformers~\cite{mot}, the streams exchange context through masked joint self-attention at each transformer layer and retain modality-specific feed-forward pathways. The initial RGB frame is encoded by the frozen video VAE
and retained as a clean latent at temporal index zero, while subsequent video latents are noised and predicted. Language and the initial tactile state provide context, allowing visual motion and contact cues to inform one another. Frozen decoders reconstruct future RGB frames and pressure maps from the predicted latent sequences.

\textbf{Anatomy-aware tactile tokenization.}
The model uses 320 taxels per hand: five $4{\times}4$ fingertip pads and one $15{\times}16$ palm pad (Fig.~\ref{fig:tactile_setup}). A dense image layout introduces artificial neighborhoods between these disconnected regions. We instead align both hands in a common anatomical coordinate system and use a shared encoder for the ten fingertips and another for the two palms. Pad identity embeddings preserve anatomical correspondence while sharing local pressure features. The resulting \emph{Split-Hands} tokenizer is pretrained with contact-weighted reconstruction, then frozen during world-model training. Weighting contact regions prevents the numerous inactive taxels from dominating the representation.

\textbf{Anchored residual tactile latents.}
Let $E_x$ and $D_x$ be the tactile encoder and decoder, and $\mathbf{z}_t=E_x(\mathbf{x}_t)$. We predict changes relative to the initial tactile state:
\begin{equation}
\mathbf{r}_t=\mathbf{z}_t-\mathbf{z}_0,\qquad
\hat{\mathbf{x}}_t=D_x(\mathbf{z}_0+\hat{\mathbf{r}}_t).
\label{eq:residual}
\end{equation}
The clean initial encoding $\mathbf{z}_0$ remains available as context. Residual prediction focuses the model on interaction-induced changes while preserving the initial pressure background.

\textbf{Dual-rate action conditioning.}
The causal video VAE preserves a separate initial latent and compresses future frames by a factor of four; tactile tokenization retains the observation rate. We therefore condition each stream at its own latent resolution:
\begin{equation}
\mathbf{c}^{v}_j=\phi_v(\mathbf{a}_{4j-3:4j}),\qquad
\mathbf{c}^{x}_t=\phi_x(\mathbf{a}_t),
\label{eq:action}
\end{equation}
for future video chunks $j$ and tactile steps $t$, with $\mathbf{c}^{v}_0=\mathbf{c}^{x}_0=\mathbf{0}$. The pose sequence uses the first-frame-relative wrist and camera representations defined in Sec.~\ref{sec:human_robot_alignment}, together with absolute hand joint angles. AdaLN-based pose conditioning injects these signals into each expert's timestep modulation. Four-frame pose chunks condition video latents, while frame-level poses preserve fine temporal cues for tactile prediction. The camera component provides egocentric motion context alongside bilateral hand motion.

\textbf{Training objective.}
We train both experts with conditional flow matching~\cite{flowmatching}:
\begin{equation}
\mathcal{L}=\mathcal{L}^{\mathrm{FM}}_{\mathrm{video}}
+\lambda_{\mathrm{tac}}\mathcal{L}^{\mathrm{FM}}_{\mathrm{tactile}}.
\label{eq:loss}
\end{equation}
The initial visual latent is excluded from the prediction loss. The tactile term predicts the residuals in Eq.~\ref{eq:residual} and is evaluated on contact or temporally changing tokens, reducing the influence of static non-contact regions. Joint attention couples the modalities without an alignment loss.

\subsection{HumanTouch Demonstration Collection}
\label{sec:method_collection}
HumanTouch records synchronized visual context, bilateral pressure, and articulated hand motion during natural manipulation. Collectors wear Moxian tactile gloves beneath Manus MetaGloves for pressure and finger tracking. HTC Vive trackers recover 6-DoF wrist poses; two wrist cameras and a head camera record complementary views. Session calibration estimates camera intrinsics, camera-to-tracker extrinsics, and a shared temporal reference. Device monitoring identifies missing or unhealthy streams.

\begin{figure}[t]
\centering
\includegraphics[width=\columnwidth]{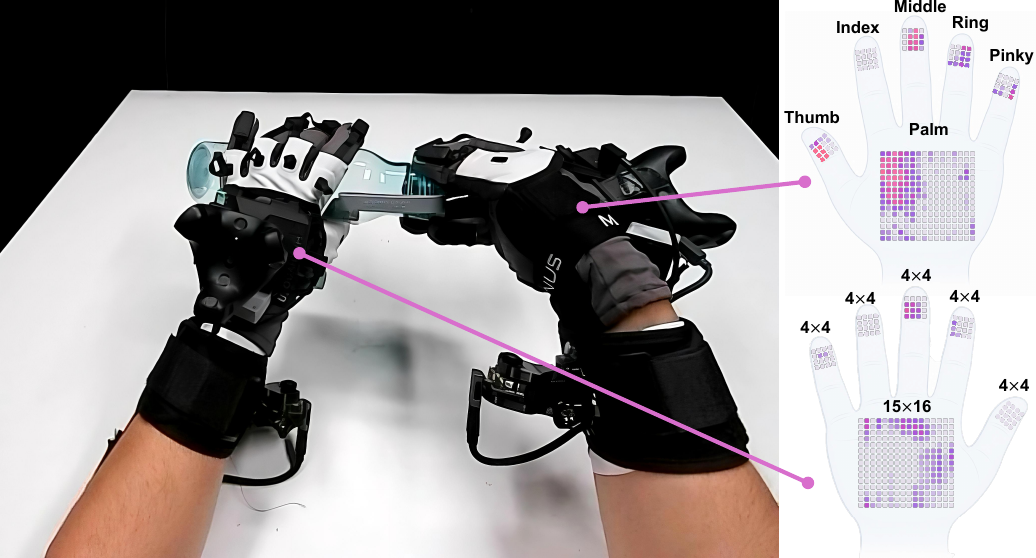}
\caption{\textbf{Wearable tactile sensing layout used in training.} Each glove records 360 taxels. The model keeps 320 of them: five fingertip pads ($4{\times}4$) and one palm pad ($15{\times}16$). The same 320-taxel layout is mounted on the robot hand, so both domains share a tactile observation space.}
\label{fig:tactile_setup}
\end{figure}

Wrist views capture local hand--object interactions, while the head camera records workspace context. Calibrated transforms register observations in a common frame, and finger tracking combined with wrist poses yields 25-keypoint 3D hand skeletons. Each episode stores RGB, tactile maps, skeletons, wrist poses, and metadata. The gloves record 360 taxels per hand at 60\,Hz; training retains the 320 sensing taxels shown in Fig.~\ref{fig:tactile_setup}. The same layout is mounted on the robot hand, giving both domains compatible tactile observations.

\subsection{Human-to-Robot Action Alignment}
\label{sec:human_robot_alignment}
Shared tactile sensing does not by itself align human and robot kinematics. We transform tracked human motion into the robot reference frame using calibrated camera--world alignment, normalize hand geometry for morphology differences, and retarget it to WujiHand2 with an inverse-kinematics mapping:
\begin{equation}
(\mathbf{e}^{h}_t,\mathbf{J}^{h}_t)
\xrightarrow{\mathcal{A}}
(\mathbf{e}^{r}_t,\mathbf{q}^{r}_t),
\label{eq:retarget}
\end{equation}
where $\mathbf{e}^{h}_t$ and $\mathbf{J}^{h}_t$ denote the human wrist pose and hand geometry; the outputs are the robot end-effector pose and 20-DoF hand configuration. This removes human-specific action coordinates while retaining the demonstrated hand motion.

Both domains then share the same retargeted hand-action representation:
\begin{equation}
\mathbf{a}_t =
[\mathbf{e}^{L}_t,\mathbf{e}^{R}_t,
 \mathbf{q}^{L}_t,\mathbf{q}^{R}_t,\mathbf{e}^{c}_t]
\in \mathbb{R}^{67}.
\label{eq:unified_action}
\end{equation}
where $\mathbf{e}$ denotes first-frame-relative wrist motion and
$\mathbf{q}\in\mathbb{R}^{20}$ denotes the hand configuration.
All streams are synchronized before training windows are constructed.
Together with the shared tactile layout, this alignment makes human
trajectories compatible with robot-world-model supervision without requiring
a learned tactile-domain mapping.
\begin{table*}[t]
\centering
\caption{
\textbf{Model design and human-to-robot scaling.}
All columns use 5h of robot data. \textnormal{Robot-only} is Split-Hands+AdaLN with 0h human data and is the origin of the scaling columns. \textnormal{+100h Human} is DexTouch-WM.
$\uparrow$ and $\downarrow$ indicate higher and lower is better, respectively.
}
\label{tab:main_eval}

\small
\renewcommand{\arraystretch}{1.25}
\setlength{\tabcolsep}{3.5pt}

\begin{tabular}{lllcccccc}
\toprule

&&&
\multicolumn{3}{c}{\textbf{Model Design}}
&
\multicolumn{3}{c}{\textbf{Human Data Scaling (5h robot fixed)}} \\

\cmidrule(lr){4-6}
\cmidrule(lr){7-9}

\textbf{Modality}
& \textbf{Capability}
& \textbf{Metric}
& \textbf{V-only}
& \textbf{Cross-Attn}
& \textbf{Robot-only}
& \textbf{+10h Human}
& \textbf{+50h Human}
& \textbf{+100h Human} \\

\midrule

\multirow{10}{*}{\rotatebox[origin=c]{90}{\textbf{Visual}}}

& \multirow{3}{*}{Pixel Fidelity}
& PSNR $\uparrow$
& 22.099 & 22.130 & \textbf{23.473}
& 24.032 & 26.874 & \textbf{27.096} \\

&
& SSIM $\uparrow$
& 0.799 & 0.801 & \textbf{0.824}
& 0.838 & 0.886 & \textbf{0.890} \\

&
& LPIPS $\downarrow$
& 0.117 & 0.118 & \textbf{0.098}
& 0.084 & \textbf{0.048} & \textbf{0.048} \\

\cmidrule(lr){2-9}

& \multirow{2}{*}{Appearance}
& Aesthetic Quality $\uparrow$
& \textbf{0.406} & 0.405 & 0.400
& 0.404 & \textbf{0.414} & \textbf{0.414} \\

&
& Image Quality $\uparrow$
& \textbf{0.718} & 0.717 & \textbf{0.718}
& 0.717 & 0.723 & \textbf{0.724} \\

\cmidrule(lr){2-9}

& \multirow{2}{*}{Representation}
& JEPA Similarity $\uparrow$
& \textbf{0.870} & 0.860 & 0.864
& 0.894 & 0.917 & \textbf{0.924} \\

&
& Subject Consistency $\uparrow$
& 0.689 & 0.681 & \textbf{0.751}
& 0.740 & 0.748 & \textbf{0.751} \\

\cmidrule(lr){2-9}

& Semantics
& Semantic Alignment $\uparrow$
& \textbf{0.829} & 0.788 & 0.813
& 0.835 & \textbf{0.866} & 0.860 \\

\cmidrule(lr){2-9}

& Action
& Trajectory Accuracy $\uparrow$
& 0.884 & 0.850 & \textbf{0.891}
& 0.900 & \textbf{0.962} & \textbf{0.962} \\

\cmidrule(lr){2-9}

& Geometry
& Geometry Error $\downarrow$
& \textbf{0.128} & 0.130 & 0.131
& 0.112 & 0.102 & \textbf{0.100} \\

\midrule

\multirow{8}{*}{\rotatebox[origin=c]{90}{\textbf{Tactile}}}

& \multirow{5}{*}{Reconstruction}
& PSNR$_{\mathrm{taxel}}$ $\uparrow$
& -- & 24.132 & \textbf{24.813}
& 24.349 & 28.401 & \textbf{29.179} \\

&
& SSIM$_{\mathrm{taxel}}$ $\uparrow$
& -- & 0.795 & \textbf{0.810}
& 0.823 & 0.876 & \textbf{0.896} \\

&
& LPIPS$_{\mathrm{heatmap}}$ $\downarrow$
& -- & 0.009 & \textbf{0.008}
& 0.008 & 0.003 & \textbf{0.002} \\

&
& MSE $\downarrow$
& -- & 0.105 & \textbf{0.093}
& 0.108 & 0.038 & \textbf{0.029} \\

&
& MAE $\downarrow$
& -- & 0.075 & \textbf{0.069}
& 0.069 & 0.040 & \textbf{0.035} \\

\cmidrule(lr){2-9}

& \multirow{3}{*}{Contact}
& Contact-MAE $\downarrow$
& -- & 0.880 & \textbf{0.806}
& 0.819 & 0.580 & \textbf{0.516} \\

&
& Contact-IoU $\uparrow$
& -- & 0.388 & \textbf{0.415}
& 0.417 & 0.534 & \textbf{0.588} \\

&
& Contact-F1 $\uparrow$
& -- & 0.523 & \textbf{0.551}
& 0.554 & 0.659 & \textbf{0.706} \\

\bottomrule
\end{tabular}
\end{table*}

\section{Experiments}
\label{sec:experiments}

We first select the architecture for joint visual--tactile prediction (Sec.~\ref{sec:exp_model_design}). With this architecture, we then ask whether scaling human data improves robot domain world model performance (Sec.~\ref{sec:exp_transfer}). Finally, we take the resulting model and examine two downstream uses: policy evaluator (Sec.~\ref{sec:exp_evaluator}) and data generator (Sec.~\ref{sec:exp_datagen}).

\subsection{Experimental Setup}
\label{sec:exp_setup}

\looseness=-1 Our experiments use egocentric human interaction data from
54 tasks and robot interaction data from 10 dexterous
manipulation tasks, covering world-model pretraining and
downstream adaptation. We investigate whether human tactile
experience transfers to robot world modeling.

\textbf{Human interaction data.}
For world-model pretraining, we collect approximately
100 hours of egocentric interaction across 50 everyday
bimanual tasks. We additionally collect human demonstrations
for four downstream tasks, forming a separate dataset
for task-specific world-model adaptation. Participants wear bilateral piezoresistive gloves with MANUS
finger trackers and VIVE wrist trackers, yielding synchronized full-hand
tactile maps, articulated finger motion, wrist 6-DoF poses, and a head-mounted
RGB view. The tasks cover transport, grasping and regrasping, impact,
insertion, articulated and deformable objects, and sustained multi-region
contact. We retarget the recorded motions to the robot action space
(Sec.~\ref{sec:human_robot_alignment}) so that these demonstrations can
supervise action-conditioned robot-world dynamics.

\textbf{Robot interaction data.}
We collect real-robot interaction on 10 dexterous manipulation tasks using a Tianji robotic arm equipped with a 20-DoF Wuji dexterous hand. Six tasks provide 5 hours of world-model pretraining data. The remaining four tasks are disjoint from the pretraining tasks and are used for downstream adaptation and evaluation. The same full-hand piezoresistive glove used for human collection is mounted on the Wuji hand, so human and robot tactile observations share one sensing interface. Each trajectory contains synchronized RGB observations, full-hand tactile measurements, and robot joint trajectories.

\textbf{Evaluation protocol.}
We evaluate visual fidelity, action consistency, tactile reconstruction, and contact dynamics. For visual prediction, we build on the multidimensional benchmarking protocol of WorldArena~\cite{shang2026worldarena} and the evaluator-oriented analysis of GigaWorld~\cite{gigaworld}. We report \emph{Aesthetic Quality} and \emph{Image Quality} for perceptual appearance; \emph{JEPA Similarity}, \emph{Semantic Alignment}, and
\emph{Subject Consistency} for representation and semantic consistency; and \emph{Trajectory Accuracy} for action-conditioned motion fidelity.
We additionally report \emph{Geometry Error} to assess geometric consistency.

For tactile prediction, we evaluate continuous signal reconstruction using PSNR$_{\mathrm{taxel}}$, SSIM$_{\mathrm{taxel}}$, LPIPS$_{\mathrm{heatmap}}$, MSE, and MAE. Since average reconstruction errors can be dominated by the large non-contact regions and therefore obscure sparse but task-critical contacts, we additionally evaluate contact dynamics using Contact-MAE, Contact-IoU and Contact-F1.

\subsection{Visuo-Tactile World Model Design}
\label{sec:exp_model_design}

This subsection uses only the five-hour robot set. We first choose a tactile tokenizer, then compare how actions are injected into a joint visual--tactile model.

\begin{figure}[t]
    \centering
    \includegraphics[width=\columnwidth]{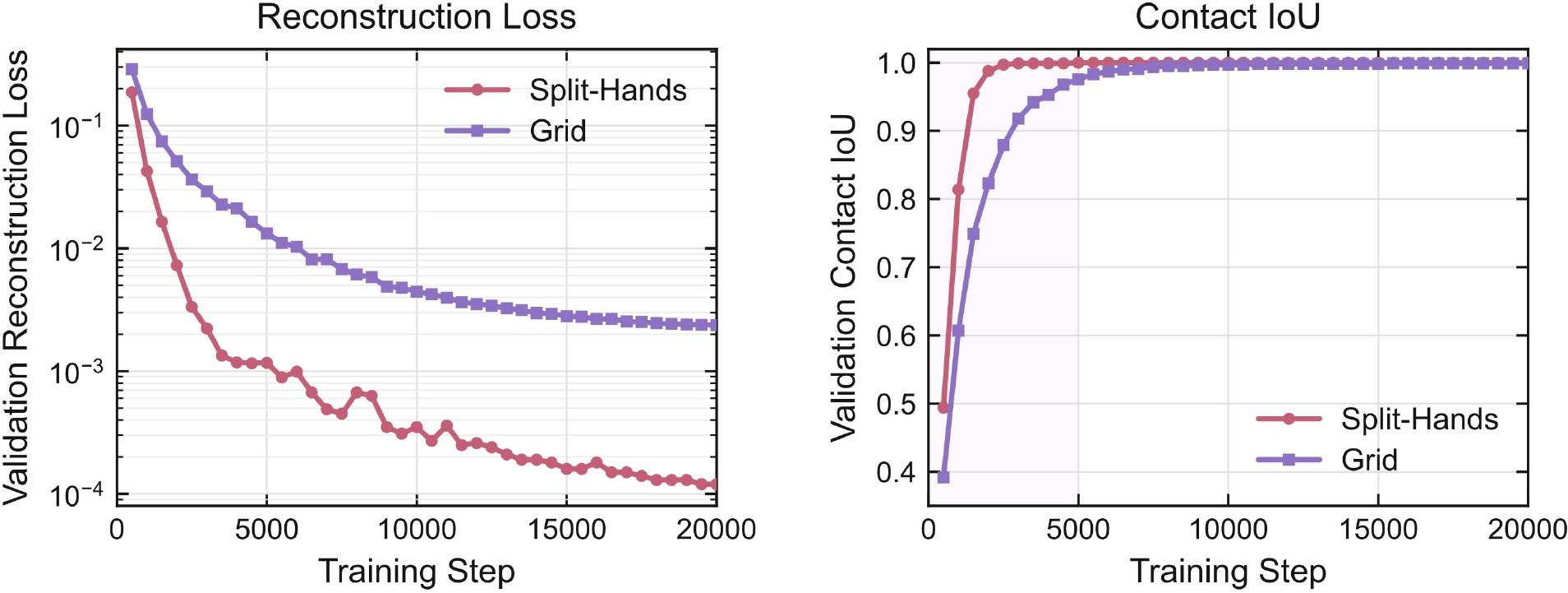}
    \caption{
    \textbf{Tactile representation ablation.}
    Validation reconstruction loss and Contact-IoU for Grid and Split-Hands.
    }
    \label{fig:tactile_tokenizer}
\end{figure}

\textbf{Anatomy-aware tactile representation.}
\looseness=-1 We first study how full-hand tactile observations should be represented.
The \emph{Grid} baseline treats the tactile map as a single spatial canvas, where the palm and fingertips are jointly encoded by a shared CNN. Although simple, this representation introduces artificial spatial
neighborhoods between anatomically disconnected regions. In contrast, our \emph{Split-Hands} representation preserves the anatomical
structure of the hand by encoding the fingertips and palms separately with shared region-specific encoders, while retaining pad identity and local position information.
As shown in Fig.~\ref{fig:tactile_tokenizer}, Split-Hands consistently achieves lower reconstruction error and substantially faster convergence in Contact-IoU than the Grid baseline. These results indicate that preserving anatomical structure provides a more
effective inductive bias for modeling sparse and localized contact patterns. We therefore use Split-Hands in all subsequent experiments.

\textbf{Action conditioning mechanism.}
We next compare three ways of injecting actions, reported in Table~\ref{tab:main_eval} as V-only, Cross-Attn, and Robot-only. V-only predicts video alone and conditions on actions by cross-attention. Cross-Attn is a joint visual--tactile model with the same cross-attention injection. Robot-only replaces that injection with AdaLN and is the architecture used in the remainder of the paper.

Introducing tactile prediction with cross-attention largely preserves
low-level visual fidelity: compared with V-only, PSNR and SSIM remain nearly unchanged and LPIPS is comparable. However, the evaluator-oriented metrics reveal a different trend. Cross-Attn reduces semantic and trajectory consistency, with Trajectory Accuracy dropping from approximately 0.88 to 0.85. This suggests that simply coupling the additional tactile stream through cross-attention does not harm pixel reconstruction, but can introduce cross-modal interference in higher-level action-conditioned dynamics.

AdaLN substantially alleviates this interference. Relative to Cross-Attn, Robot-only improves visual PSNR from 22.1 to 23.5 and LPIPS from 0.12 to 0.10, while recovering and slightly surpassing V-only in Trajectory Accuracy (0.89 vs.\ 0.88). The improvement is even more consistent in the tactile domain: Contact-IoU increases from 0.39 to 0.42 and Contact-F1 from 0.52 to 0.55.
Together, these results indicate that AdaLN provides a more effective
action-conditioning mechanism for jointly modeling visual motion and contact dynamics. We therefore use the Robot-only architecture (Split-Hands + AdaLN) for all subsequent experiments.

% \vspace{-3mm}
\begin{table}[t]
\centering
\caption{
\textbf{Policy evaluation in the real world and in two imagined environments.}
The same policies are rolled out in the real world, in \emph{WM-Robot} (DexTouch-WM adapted on 200 robot trajectories), and in \emph{WM-Mix} (adapted on 100 robot + 100 human trajectories).
}
\label{tab:policy_evaluator}

\scriptsize
\renewcommand{\arraystretch}{1.08}
\setlength{\tabcolsep}{3.0pt}

\begin{tabular}{llcccc}
\toprule
\textbf{Policy}
& \textbf{Eval.}
& \textbf{Place Shoes}
& \textbf{Place Phone}
& \textbf{Stack Bowls}
& \textbf{Stand Bottle} \\
\midrule

\multirow{3}{*}{FTP-1}
& Real world  & 0.725 & 0.800 & 0.500 & 0.900 \\
& WM-Robot  & 0.975 & 0.975 & 0.750 & 0.900 \\
& WM-Mix & 0.963 & 0.975 & 0.800 & 1.000 \\
\midrule

\multirow{3}{*}{$\pi_{0.5}$}
& Real world  & 0.450 & 0.700 & 0.850 & 0.500 \\
& WM-Robot  & 0.950 & 0.925 & 0.950 & 0.850 \\
& WM-Mix & 0.900 & 0.925 & 0.950 & 0.800 \\
\midrule

\multirow{3}{*}{X-VLA}
& Real world  & 0.500 & 0.600 & 0.300 & 0.600 \\
& WM-Robot  & 0.600 & 0.725 & 0.150 & 0.150 \\
& WM-Mix & 0.925 & 0.700 & 0.250 & 0.400 \\
\bottomrule
\end{tabular}
\end{table}

\begin{table}[t]
\centering
\caption{Pearson correlation ($\uparrow$) and MMRV ($\downarrow$) between world-model and real-world evaluation.}
\label{tab:wm_real_corr}

\footnotesize
\renewcommand{\arraystretch}{1.08}
\setlength{\tabcolsep}{2.5pt}

\begin{tabular}{@{}llcccc@{}}
\toprule
\textbf{Model} & \textbf{Metric}
& \shorttask{Place}{Shoes}
& \shorttask{Place}{Phone}
& \shorttask{Stack}{Bowls}
& \shorttask{Stand}{Bottle} \\
\midrule
\multirow{2}{*}{WM-Robot}
& Pearson
& 0.400 & \textbf{0.945} & \textbf{0.906} & 0.334 \\
& MMRV
& 0.033 & 0 & 0 & 0.067 \\
\midrule
\multirow{2}{*}{WM-Mix}
& Pearson
& \textbf{0.972} & 0.939 & 0.889 & \textbf{0.577} \\
& MMRV
& \textbf{0} & 0 & 0 & 0.067 \\
\bottomrule
\end{tabular}
\end{table}

\begin{figure*}[t]
\centering
\includegraphics[width=\textwidth]{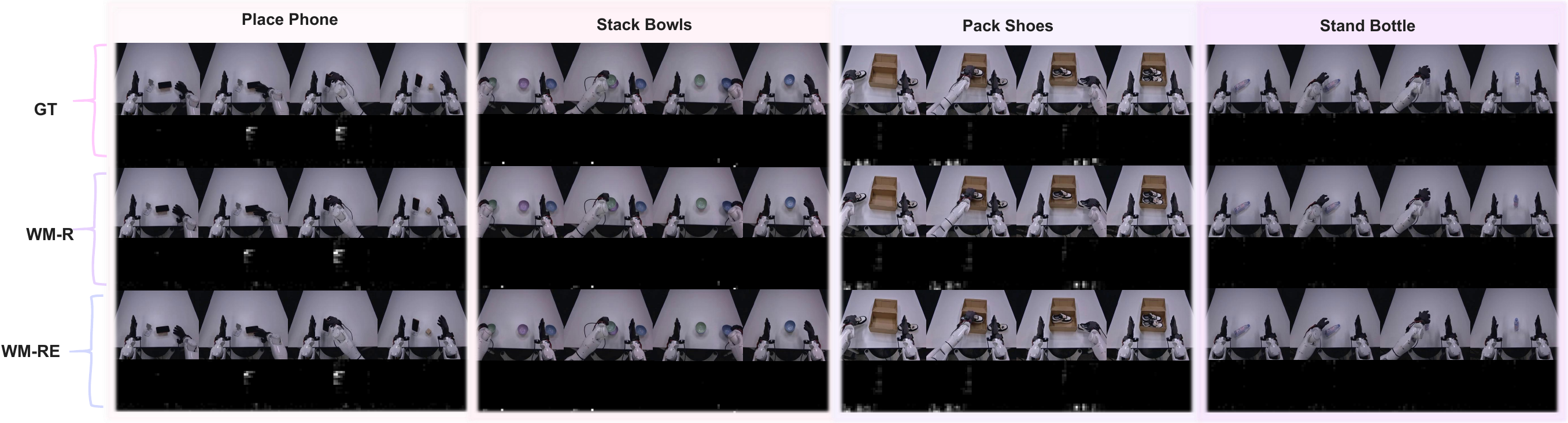}
\caption{\textbf{Generated visuo-tactile demonstrations.} Columns are the four downstream tasks. Rows compare a held-out real episode (GT) with action-conditioned rollouts from WM-Robot and WM-Mix. Each cell shows synchronized RGB frames and tactile maps from one trajectory. These clips are the synthetic demonstrations used for policy training, not in-model policy evaluation.}
\label{fig:gen_data_wm}
\end{figure*}

\subsection{Human-to-Robot Transfer}
\label{sec:exp_transfer}

We next ask whether human interaction scales robot-domain prediction when the robot training budget is held fixed. Starting from Robot-only, we keep the five-hour robot set unchanged and add 10, 50, or 100 hours of human data. These models appear in Table~\ref{tab:main_eval} as +10h Human, +50h Human, and +100h Human; the last of these is \method{} and initializes task-level adaptation below.
All variants are evaluated on the same held-out robot episodes from six manipulation tasks. Human and robot task sets are disjoint, so improvements indicate cross-task transfer rather than task-specific augmentation.

Table~\ref{tab:main_eval} shows a clear scaling trend.
From Robot-only to +100h Human, visual PSNR improves from 23.47 to 27.10, Trajectory Accuracy from 0.89 to 0.96, and Geometry Error from 0.13 to 0.10.
The gains cover pixel fidelity, representation, motion, and geometry, while aesthetic and image-quality scores stay nearly unchanged. Human data therefore contributes transferable interaction structure.

Contact transfer is stronger. At +100h Human, tactile PSNR improves from 24.81 to 29.18, Contact-IoU from 0.42 to 0.59, and Contact-F1 from 0.55 to 0.71. +10h Human is almost flat on tactile metrics; the main gains appear at 50--100h. Visual metrics saturate between +50h and +100h Human, while contact prediction continues to improve. Aligned human interaction is therefore a complementary scaling axis: \method{} improves held-out robot visual and contact prediction without collecting more robot data.

\subsection{World Models as Policy Evaluators}
\label{sec:exp_evaluator}

\textbf{Adaptation and evaluation protocol.}
We evaluate FTP-1~\cite{yuan2026ftp}, $\pi_{0.5}$~\cite{intelligence2025pi05}, and
X-VLA~\cite{zheng2025xvla} on four tasks (Table~\ref{tab:policy_evaluator}). Each task contains 300 robot trajectories split into disjoint subsets $A$, $B$, and $C$ (100 each). From the same \method{} checkpoint, WM-Robot adapts on 200 robot trajectories from $B$ and $C$, whereas WM-Mix uses 100 robot trajectories from $B$ and 100 task-specific human demonstrations disjoint from the HumanTouch scaling corpus (Sec.~\ref{sec:exp_transfer}). All policies train on the same 200 real trajectories from $A$ and $B$ and are evaluated on the robot and in both world models using matched initial frames. World-model rollouts are closed-loop, with predicted observations conditioning subsequent policy actions. Raw task scores assign one point per shoe pickup or placement (Place Shoes; max 4), per stacked bowl (Stack Bowls; max 2), for phone pickup and placement on the stand separately (Place Phone; max 2), and for successful upright bottle placement (Stand Bottle; max 1). Imagined rollouts incur a 0.5-point penalty for clear physical inconsistencies. All scores are then mapped to $[0,1]$ under the task-specific rubrics, with $1$ denoting full task completion. Each policy--task pair is evaluated with 10 matched rollouts per environment, each independently scored by five human raters. Reported scores average across raters and rollouts.

\textbf{Evaluation metrics.}
As shown in Table~\ref{tab:wm_real_corr}, we assess linear association between imagined and real-world scores using Pearson correlation, and ranking consistency using mean maximum rank violation (MMRV)~\cite{li2024simpler,jain2025polaris}. For task $k$, let $R_{ik}$ and $S_{ik}$ denote the real and imagined scores of policy $i$. We compute $r_k=\mathrm{corr}_i(R_{ik},S_{ik})$ and define $v_{ij,k}=1$ when $(R_{ik}-R_{jk})(S_{ik}-S_{jk})<0$, and zero otherwise. Then
\begin{equation}
\mathrm{MMRV}_k=\frac{1}{N}\sum_{i=1}^{N}\max_j
\bigl(|R_{ik}-R_{jk}|\,v_{ij,k}\bigr),
\label{eq:mmrv}
\end{equation}
where $N=3$. MMRV penalizes each policy's largest rank inversion by the corresponding real-score gap; lower is better.

\textbf{Agreement with real-world evaluation.}
Table~\ref{tab:wm_real_corr} shows that WM-Mix achieves higher mean Pearson correlation than WM-Robot (0.844 versus 0.646) and lower mean MMRV (0.017 versus 0.025). On Place Phone and Stack Bowls, both models preserve the real-world policy ordering, with zero MMRV, although WM-Robot has slightly higher Pearson correlation. On Place Shoes, WM-Mix corrects the rank inversion between $\pi_{0.5}$ and X-VLA present in WM-Robot. Stand Bottle remains challenging: Pearson increases from 0.334 to 0.577, but both models reverse the order of $\pi_{0.5}$ and X-VLA, yielding the same MMRV of 0.0667. Thus, the gains improve average evaluation fidelity without resolving every task's ranking errors.

\textbf{Human-data substitution.}
The 12 paired scores in Table~\ref{tab:policy_evaluator} give a pooled Pearson correlation of $r=0.925$ between WM-Robot and WM-Mix. This measures consistency between adaptation variants; Table~\ref{tab:wm_real_corr} assesses each variant against real-world evaluation. Together, the results support replacing half of the task-specific robot adaptation demonstrations with human demonstrations while retaining evaluator consistency and improving average alignment with real-world evaluation.

% \textbf{Scope of the evidence.}
% Table~\ref{tab:policy_evaluator} also shows that score calibration remains imperfect: mean absolute errors against real-world scores are approximately 0.223 for WM-Robot and 0.222 for WM-Mix. The improvement therefore concerns relative policy assessment, while accurate absolute scores remain an open limitation. With only three policies per task, the correlations provide descriptive evidence within the evaluated setting.

% \vspace{-2mm}
\begin{table}[t]
\centering
\caption{
\textbf{Policy learning on real versus half-imagined data.}
\emph{Real} trains the policy on 200 real trajectories. \emph{+WM-Robot} and \emph{+WM-Mix} train it on 100 real trajectories plus 100 imagined trajectories from that world model.
}
\label{tab:data_generator}

\scriptsize
\renewcommand{\arraystretch}{1.08}
\setlength{\tabcolsep}{3.0pt}

\begin{tabular}{llcccc}
\toprule
\textbf{Policy}
& \textbf{Train Data}
& \textbf{Place Shoes}
& \textbf{Place Phone}
& \textbf{Stack Bowls}
& \textbf{Stand Bottle} \\
\midrule

\multirow{3}{*}{FTP-1}
& Real   & 0.725 & 0.800 & 0.500 & 0.900 \\
& +WM-Robot  & 0.750 & 0.850 & 0.350 & 0.800 \\
& +WM-Mix & 0.775 & 0.700 & 0.500 & 0.800 \\
\midrule

\multirow{3}{*}{$\pi_{0.5}$}
& Real   & 0.450 & 0.700 & 0.850 & 0.500 \\
& +WM-Robot  & 0.400 & 0.550 & 0.800 & 0.500 \\
& +WM-Mix & 0.375 & 0.350 & 0.850 & 0.400 \\
\midrule

\multirow{3}{*}{X-VLA}
& Real   & 0.500 & 0.600 & 0.300 & 0.600 \\
& +WM-Robot  & 0.675 & 0.300 & 0.450 & 0.600 \\
& +WM-Mix & 0.550 & 0.600 & 0.450 & 0.000 \\
\bottomrule
\end{tabular}
\end{table}

\begin{table}[t]
\centering
\caption{Pearson correlation ($\uparrow$) and MMRV ($\downarrow$) between policies trained with world-model-generated data and policies trained with real data across tasks.}
\label{tab:wm_data_generator}

\scriptsize
\renewcommand{\arraystretch}{1.08}
\setlength{\tabcolsep}{3.0pt}

\begin{tabular}{llcccc}
\toprule
\textbf{Data} & \textbf{Metric} 
& \textbf{Place Shoes} 
& \textbf{Place Phone} 
& \textbf{Stack Bowls} 
& \textbf{Stand Bottle} \\
\midrule

\multirow{2}{*}{+WM-Robot}
& Pearson $\uparrow$ 
& 0.783 & \textbf{0.999} & 0.836 & \textbf{0.996} \\
& MMRV $\downarrow$ 
& 0 & \textbf{0} & 0.133 & \textbf{0} \\
\midrule

\multirow{2}{*}{+WM-Mix}
& Pearson $\uparrow$ 
& \textbf{0.961} & 0.277 & \textbf{0.968} & 0.721 \\
& MMRV $\downarrow$ 
& 0 & 0.067 & \textbf{0} & 0.067 \\

\bottomrule
\end{tabular}
\end{table}
% \vspace{-1mm}

\subsection{World Models as Data Generators}
\label{sec:exp_datagen}
We next test whether imagined trajectories support policy learning on the physical robot. For each task, policies in the Real setting are trained on
200 real trajectories from splits $A$ and $B$.
The +WM-Robot and +WM-Mix settings each use 100 real
trajectories from $B$ and 100 synthetic trajectories generated
by the corresponding world model. Each synthetic trajectory
retains the initial observations and recorded action sequence
of a trajectory in $A$, while its future RGB and tactile
observations are predicted by the world model.
Split $A$ is excluded from all world-model training. All final policy scores in Table~\ref{tab:data_generator} are measured on the robot.

Figure~\ref{fig:gen_data_wm} visualizes the generated demonstrations. Both adapted models produce action-conditioned rollouts whose object motion and contact timing follow the held-out real episode. We use these trajectories as the synthetic half of the +WM-Robot and +WM-Mix training sets.

The effect depends on the policy. FTP-1 trained with +WM-Mix achieves a four-task mean of 0.694 versus 0.731 for all-real training. For $\pi_{0.5}$, the mean decreases from 0.625 to 0.494; for X-VLA, it falls from 0.500 to 0.400, including a zero score on Stand Bottle. +WM-Robot yields means of 0.688, 0.563, and 0.506, respectively. Human-adapted world models can therefore supply useful demonstrations, but their evaluator consistency does not ensure equivalent policy-training utility. The usefulness of synthetic data therefore depends on the policy and task, even when the generated rollouts resemble real episodes and the source models behave similarly as evaluators.

\section{Conclusion}
\label{sec:conclusion}

We presented \method{}, an action-conditioned visuo-tactile world model that learns from human and robot interaction through a shared tactile interface and aligned actions. With robot training data fixed at 5\,h, scaling human interaction to 100\,h improves visual, geometric, and contact prediction on unseen robot episodes. The adapted model further serves as a policy evaluator and as a source of synthetic demonstrations for policy training. These findings support human interaction as a complementary source of supervision for dexterous robot world models.
\newpage
% \section*{ACKNOWLEDGMENT}

% We thank D-Robotics for their 
\bibliographystyle{abbrv}
\bibliography{IEEEfull,references}

\end{document}